\documentclass[10pt,twocolumn,letterpaper]{article}

\usepackage{cvpr}              
\definecolor{cvprblue}{rgb}{0.21,0.49,0.74}
\usepackage[pagebackref,breaklinks,colorlinks,allcolors=cvprblue]{hyperref}
\usepackage{algorithmic}
\usepackage{multirow}
\usepackage{bbding}
\usepackage{subcaption}
\usepackage{caption}
\usepackage{color} 
\usepackage{xcolor}
\usepackage{graphicx}
\usepackage{epigraph} 
\definecolor{darkgreen}{rgb}{0,0.4,0}
\definecolor{lightgreen}{rgb}{0.89,0.94,0.85}
\definecolor{darkdarkgreen}{rgb}{0.26,0.40,0.16}

\usepackage[linesnumbered,ruled,vlined]{algorithm2e}
\SetKwBlock{Loop}{loop}{end loop}
\SetKwInput{KwInput}{Require}                

\SetCommentSty{mycommfont}
\def\algcomment#1{\textcolor[rgb]{0,0.6,0}{\# #1}}

\def\paperID{} 
\def\confName{CVPR}
\def\confYear{2026}
\newcommand{\ourmethod}{\textsc{RayNova}\xspace}

\title{\texorpdfstring{\raisebox{-0.3em}{\includegraphics[height=1.3em]{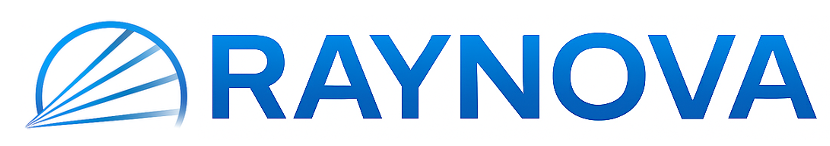}}}{\ourmethod}: Scale-Temporal Autoregressive World Modeling in Ray Space}

\author{Yichen Xie$^{1,2}$\thanks{Equal Contribution. Work done during internship at Applied Intuition.}, Chensheng Peng$^{1,2,*}$, Mazen Abdelfattah$^{1}$, Yihan Hu$^{1}$, Jiezhi Yang$^{1}$, Eric Higgins$^{1}$\\Ryan Brigden$^{1}$, Masayoshi Tomizuka$^{2}$, Wei Zhan$^{1,2}$\thanks{Correspondence: wei.zhan@applied.co}\\
$^1$ Applied Intuition, $^2$ UC Berkeley
}

\begin{document}
\maketitle

\begin{abstract}
World foundation models aim to simulate the evolution of the real world with physically plausible behavior. Unlike prior methods that handle spatial and temporal correlations separately, we propose \ourmethod{}, a geometry-agonistic multiview world model for driving scenarios that employs a dual-causal autoregressive framework. It follows both scale-wise and temporal topological orders in the autoregressive process, and leverages global attention for unified 4D spatio-temporal reasoning. Different from existing works that impose strong 3D geometric priors, \ourmethod{} constructs an isotropic spatio-temporal representation across views, frames, and scales based on relative Plücker-ray positional encoding, enabling robust generalization to diverse camera setups and ego motions. We further introduce a recurrent training paradigm to alleviate distribution drift in long-horizon video generation. \ourmethod{} achieves state-of-the-art multi-view video generation results on nuScenes, while offering higher throughput and strong controllability under diverse input conditions, generalizing to novel views and camera configurations without explicit 3D scene representation. Our code will be released at \href{https://raynova-ai.github.io/}{https://raynova-ai.github.io/}.

\end{abstract}    
\section{Introduction}
\label{sec:intro}

\begin{figure*}
    \centering
    \begin{subfigure}{\linewidth}
        \includegraphics[width=\linewidth]{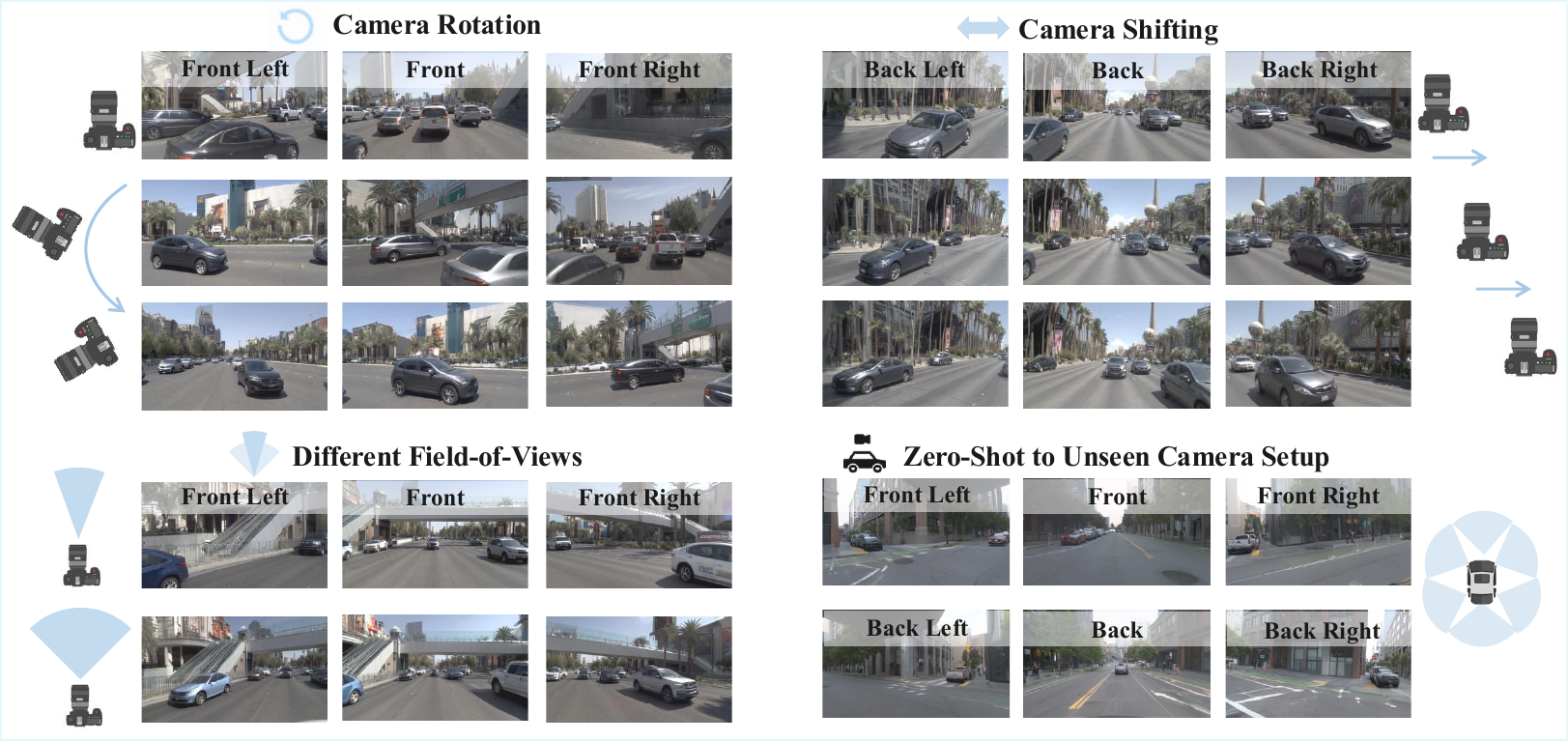}
        \caption{\ourmethod{} is not bound to any specific sensor setups, allowing flexible video generation in continuous 4D world space.}
        \label{fig:teaser_1}
    \end{subfigure}
    \begin{subfigure}{\linewidth}
        \includegraphics[width=\linewidth]{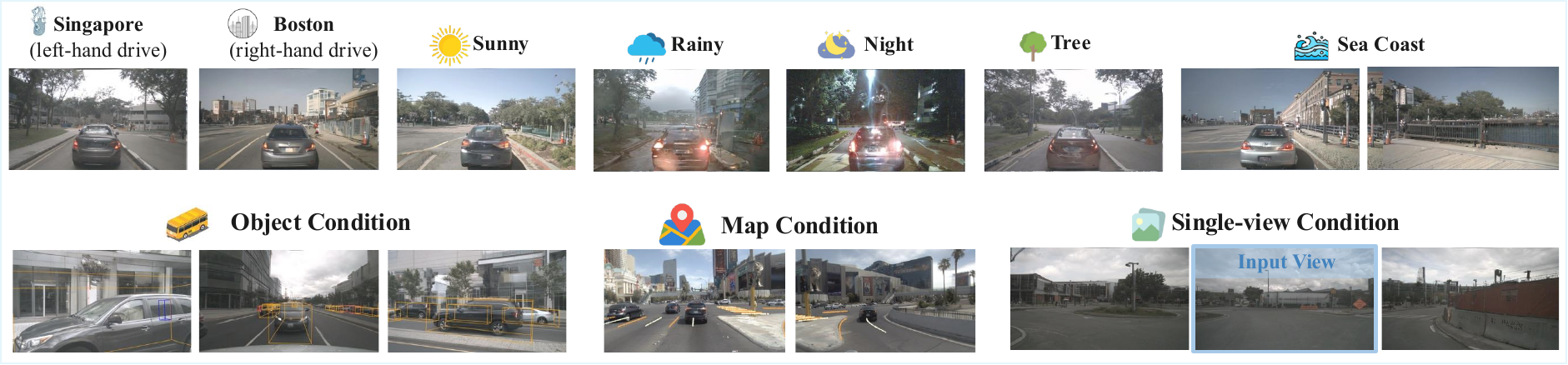}
        \caption{\ourmethod{} takes diverse input conditions with high fidelity.}
    \end{subfigure}
    \begin{subfigure}{\linewidth}
        \includegraphics[width=\linewidth]{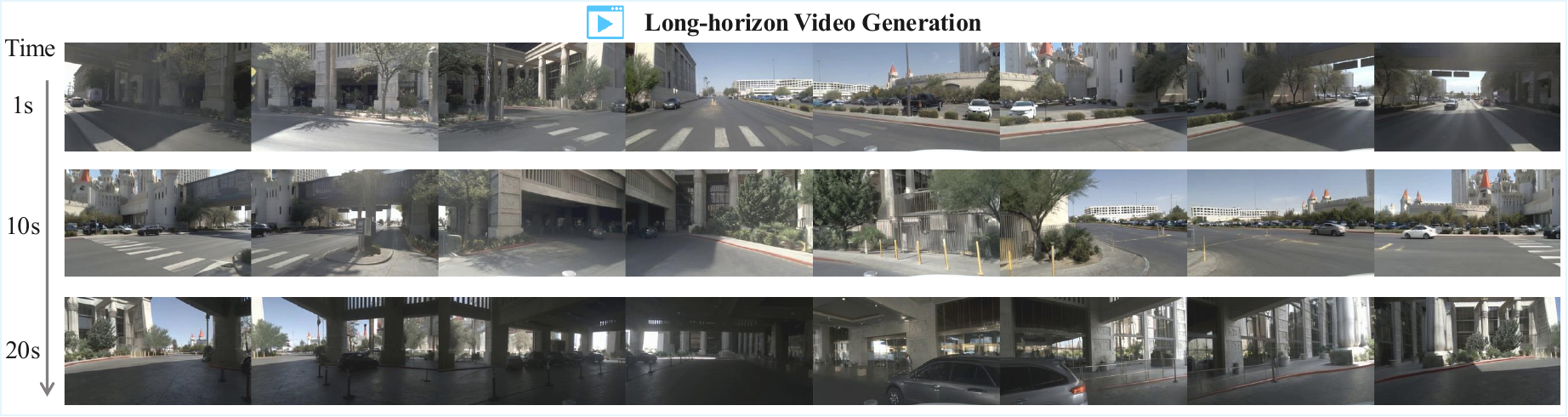}
        \caption{\ourmethod{} can generate long-horizon multi-view videos in a large range.}
    \end{subfigure}
    \vspace{-20pt}
    \caption{Demonstrations of \ourmethod{} as Versatile World Foundation Model.}
    \label{fig:teaser}
    \vspace{-10pt}
\end{figure*}

\epigraph{Space by itself, and time by itself, are doomed to fade away into mere shadows, and only a kind of union of the two will preserve an independent reality.}{--- \textit{Hermann Minkowski}}

World foundation models (WFMs)~\cite{lecun2022path,openai2024sora,bruce2024genie,ha2018world, wfm2024nvidia} aim to simulate the evolution of complex scenes within a dynamic real world governed by physical laws. In a continuous 4D spatio-temporal world, a world model is designed to generate physically plausible images or videos captured by cameras positioned at specific locations and timesteps.

To impose physical plausibility, prior work commonly considers spatial and temporal constraints separately. For spatial reasoning, some methods leverage positional correlations across multiple viewpoints, including camera adjacency~\cite{gao2023magicdrive,wen2024panacea} and cross-view projection~\cite{guo2025dist4ddisentangledspatiotemporaldiffusion}. For temporal coherence, previous works assume strong multi-frame dependencies among images captured by the same cameras~\cite{wen2024panacea,wang2024driving,russell2025gaia2controllablemultiviewgenerative} and thereby exploit existing techniques from video generation, such as temporal VAEs~\cite{yang2025cogvideoxtexttovideodiffusionmodels,brooks2024video}. However, this decoupled design fundamentally restricts the model's flexibility to handle novel sensor configurations and rapid camera motions. Alternatively, some approaches construct explicit 3D scene representations (\eg point clouds~\cite{xie2024x,zhou2025hermes} or BEV/volume features~\cite{zhang2024bevworld,lu2023wovogen}), within a unified global coordinate system to enhance spatio-temporal consistency. Although empirically effective within constrained domains, this geometry-forcing approach impedes generalization to unconstrained open-world environments beyond the distribution of training scenarios.

To overcome these limitations, we pursue a universal framework that retains physical plausibility with minimal inductive biases. We propose \ourmethod{}, a world foundation model supported by a \textit{dual-causal autoregressive strategy} over both scale and time. The scale-wise autoregression is built up on ``next-scale prediction" visual autoregressive model~\cite{tian2024visual,han2024infinity}, enabling the model to capture intricate visual patterns across multiple levels of abstraction. The temporal autoregression works frame-by-frame in the unified 4D spatio-temporal space to improve temporal coherence.

 The core insight of \ourmethod{} lies in a novel spatio-temporal conditioning mechanism that represents the physical world with \textit{relative positional encoding in camera ray space}. This ray-level representation is isotropic in continuous 4D space, thereby reducing dependency on specific camera configurations, motion patterns, or view overlaps. By encoding relative rather than absolute positions, it naturally supports extrapolation beyond the training distribution and generalizes to unconstrained real-world conditions. Leveraging this representation, \ourmethod{} performs coherent reasoning across multi-view and multi-frame images, accommodating diverse input conditions within a unified 4D geometric space. We highlight the following aspects of our method:
\begin{itemize}
    \item \textbf{Versatile World Foundation Model.} Our method supports diverse input and output formats for various use cases with a single model. Video generation can be conditioned on multiple optional control signals (\eg, texts, objects, maps, images), while output formats remain flexible in terms of perspectives, resolutions, and frame rates.
    \item \textbf{Scalable Data-Driven Framework.} Our training paradigm is capable of ingesting heterogeneous training data from diverse sources with different sensor configurations. Unlike prior works with handcrafted biases~\cite{guo2025dist4ddisentangledspatiotemporaldiffusion,chen2025geodrive}, \ourmethod{} does not require auxiliary supervision such as depth maps, point clouds, or optical flow.
    \item \textbf{Extendable Position Embedding.} Our relative ray-level positional encoding supports extrapolation beyond the training range, theoretically allowing an unbounded spatial extent. We also develop a recurrent training paradigm to mitigate distribution drift in long video generation.
    \item \textbf{Efficient Video Generation.} Our hierarchical multi-scale representation enables rapid progression from coarse abstractions to fine-grained details, and can be seamlessly integrated with existing acceleration techniques for visual autoregressive models~\cite{guo2025fastvar}.
\end{itemize}

We evaluate \ourmethod{} on driving datasets, demonstrating superior performance in both video generation and novel view synthesis. The results demonstrate high visual realism, strong spatio-temporal coherence, and high fidelity to multiple control signals together with substantial computational efficiency. These advantages enhance its practicality for physical-world applications such as autonomous driving simulation.
\section{Related Works}

\paragraph{Video Generative Models.}
Recent advances in generative models have pushed the fidelity and diversity of monocular video synthesis to new heights. Early transformer-based, masked autoregressive approaches operate on discrete tokens to generate high-quality frames in sequence~\cite{yan2021videogpt, micheli2022transformers, ge2022long, villegas22phenaki, yu2023magvit}. In parallel, diffusion-based pipelines iteratively denoise latent or pixel representations, yielding vivid motion and appearance~\cite{blattmann2023stable, openai2024sora, agarwal2025cosmos, polyak2024movie, runway2024gen3, ge2024preservecorrelationnoiseprior, ma2025lattelatentdiffusiontransformer, yang2025cogvideoxtexttovideodiffusionmodels}. In just a few years, these models excel at visual quality, offering realistic and vivid generated videos. Most of these approaches focus on text-to-video tasks, generating videos conditioned on textual prompts~\cite{ge2024preservecorrelationnoiseprior, ma2025lattelatentdiffusiontransformer, yang2025cogvideoxtexttovideodiffusionmodels}. Some other models also generate videos conditioned on reference images~\cite{wang2024unianimatetamingunifiedvideo, ren2024consisti2venhancingvisualconsistency}, or reference videos~\cite{liu2024stablev2vstablizingshapeconsistency}. In this paper, we go beyond video generation for a versatile world foundation model with flexible multi-view camera setups, ego motions, and input conditions.

\paragraph{3D Scene Representation Learning.} To more accurately simulate the geometry and dynamics of real-world scenes, several approaches leverage explicit 3D priors to enhance the geometric consistency, typified by NeRF~\cite{mildenhall2021nerf, kerbl20233d} and Gaussian splatting~\cite{chen2025omnireomniurbanscene, yan2024streetgaussiansmodelingdynamic, peng2024desire,zhou2024drivinggaussiancompositegaussiansplatting} variants. Despite high-quality image renderings, their inherent inductive biases limit the adaptability of models to complex scenarios beyond the training distribution. On the other hand, several approaches work on the geometry-free representation with improved generalization ability, most of which resort to camera ray space~\cite{sajjadi2022object,hong2023lrm,jin2024lvsm,jiang2025rayzer}. However, despite these advances, they still rely on the camera rays in the global coordinate, which serves as another bottleneck of generalization for extrapolation beyond the training distribution. In contrast, we exploit the relative positions in the camera ray space for spatio-temporal grounding with minimal inductive biases, which boosts the video generation in a large range.

\paragraph{World Models for Autonomy.}
Autonomous driving serves as an ideal testbed for physically grounded world models due to its stringent demands for modeling complex geometric environments, dynamic actor interactions, and the need for coherent spatio-temporal representations. Early efforts~\cite{wang2023drivedreamerrealworlddrivenworldmodels,hu2023gaia1generativeworldmodel} demonstrate action-conditioned video generation but are constrained primarily to a single front-view perspective. Later studies~\cite{wang2023drivingfuturemultiviewvisual,wang2023drivedreamer,wen2024panacea} have extended video generation to multi-view cameras with diverse input conditions such as textual descriptions, object bounding boxes, and HD maps. To enhance geometric consistency and structural controllability, the community commonly introduces various inductive biases. Several approaches~\cite{wang2023drivingfuturemultiviewvisual,gao2023magicdrive,wen2024panacea} count the spatial consistency of the mutual condition between adjacent cameras, but this design binds the model to some fixed camera configurations. Other efforts incorporate explicit 3D scene representations such as point clouds~\cite{xie2024x,zhou2025hermes,zhang2024bevworld}, occupancy grids~\cite{lu2023wovogen}, 3D latent features~\cite{zuo2025gaussianworld,gao2024magicdrive3d} to ensure spatial alignment in world space. However, the reliability of the above 3D structural priors highly depends on the overlap between camera perspective views. These representations also limit the world models to a constrained 3D space, hindering generalization in the open world. In contrast, our approach uniquely learns a continuous 4D latent representation based on the relative position in the camera ray space that jointly encodes spatial and temporal dimensions, enforcing multi-view spatial consistency and frame-to-frame temporal coherence with minimal hand-crafted biases.

\section{Methodology}
\begin{figure*}[htb]
    \centering
    \includegraphics[width=\linewidth]{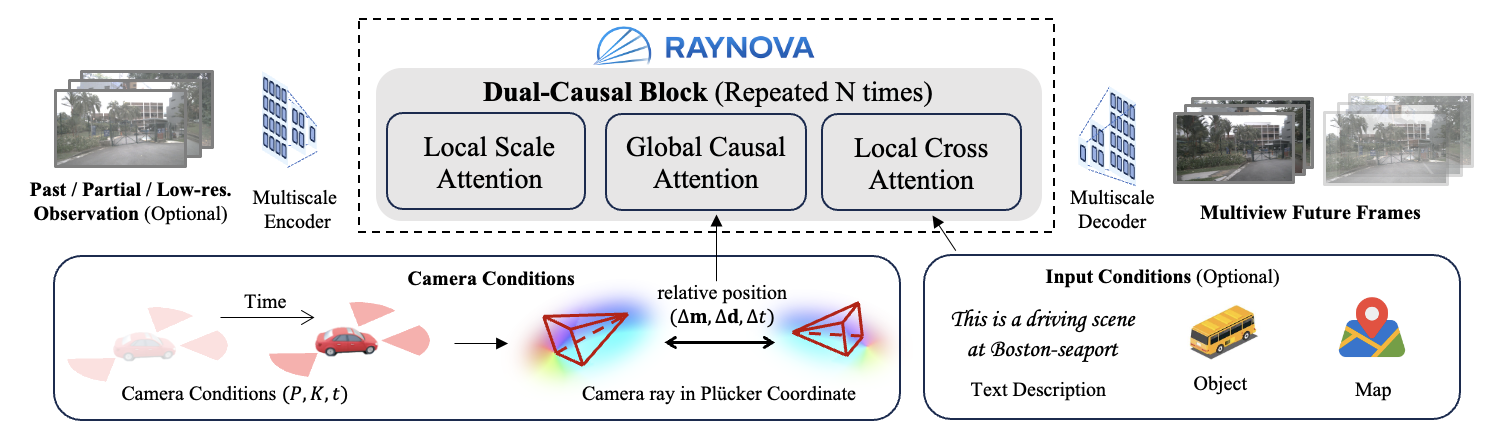}
    \vspace{-20pt}
    \caption{Overview of \ourmethod{} Framework. \ourmethod{} is composed of dual-casual (scale and time) blocks. The local scale attention and local cross attention works on each image indepedently, while the global causal attention works across multi-view and multiframe images enhanced with a unified ray-level relative position embedding for better spatio-temporal consistency.}
    \label{fig:overview}
\end{figure*}

\ourmethod{}, as shown in Fig.~\ref{fig:overview}, is a versatile world foundation model with autoregression. After introducing preliminary ``next-scale prediction" work (Sec.~\ref{sec:preliminary}), we formulate the dual-causal autoregressive framework for multi-view video generation in Sec.~\ref{sec:formulation}. Critical for coherent spatio-temporal reasoning, we construct an isotropic 4D position embedding based on the relative positions between camera rays in Sec.~\ref{sec:space}. This enables the development of a transformer-based geometry-free framework with minimal inductive biases for world modeling that takes diverse control signals in Sec.~\ref{sec:framework}. Finally, to alleviate the distribution drift in the autoregressive generation of long videos, we develop a recurrent training paradigm that aligns the distributions in the training and inference stages (Sec.~\ref{sec:training}).

\subsection{Preliminary: Next-Scale Prediction}
\label{sec:preliminary}
Unlike vanilla autoregressive models~\cite{lee2022autoregressive,yu2021vector} that flatten the 2D grids of images into 1D tokens, some recent work~\cite{tian2024visual,han2024infinity} shifts from ``next-token prediction” to ``next-scale prediction” strategy. Each image is quantized by the tokenizer into $K$ multiscale token maps $X_{1:K}=(X_1, X_2,\dots,X_K)$ with increasingly higher resolutions $h_k\times w_k, k=1,2,\dots,K$. The autoregressive likelihood is formulated as follows.
\vspace{-5pt}
\begin{equation}
    p(X_1,X_2,\dots,X_K)=\prod_{k=1}^Kp(X_k|X_1,X_2,\dots,X_{k-1})
    \label{eq:var}
\end{equation}
where $X_k$ is the token map at scale $k$ containing $h_k\times w_k$ visual tokens. The sequence $(X_1, \dots, X_{k-1})$ is the prefix context for the prediction of $X_k$. During the $k$-th autoregressive step, all $h_k\times w_k$ tokens are generated in parallel.

\begin{figure}[tb]
    \centering
    \includegraphics[width=0.9\linewidth]{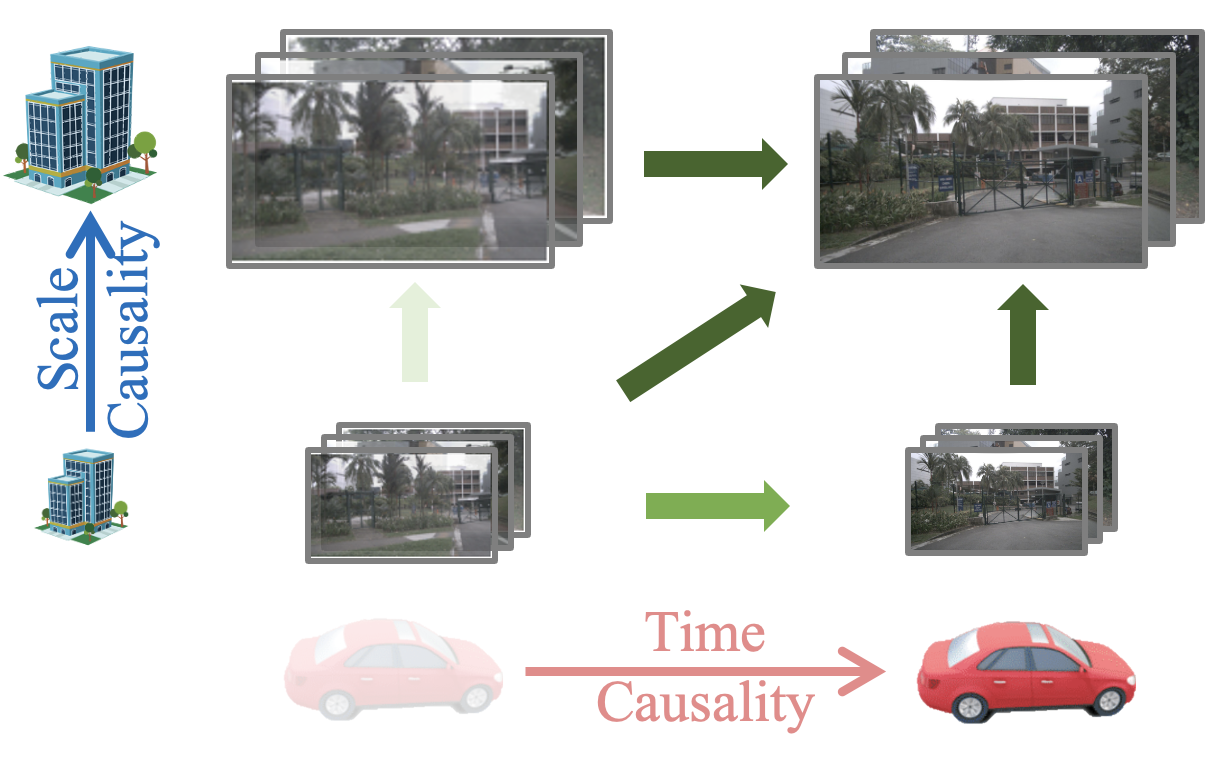}
    \vspace{-10pt}
    \caption{Dual-Causality for Multi-View Video Generation. \textcolor{darkgreen}{Green arrows} represent the causal dependency, while the darkness indicates the topological order of autoregression (from \textcolor{lightgreen}{light} to \textcolor{darkdarkgreen}{dark}).}
    \label{fig:dual-causal}
    \vspace{-5pt}
\end{figure}

\subsection{Formulation of Dual-Causal Autoregression}
\label{sec:formulation}
We formulate our proposed \ourmethod{} for world modeling via multi-view videos beyond image-level ``next-scale prediction". In favor of versatility, each image $I^{v,t}$ is tokenized independently as a multiscale token map $X^{v,t}_{1:K}$ for each frame $t=1,\dots,T$ and each camera view $v=1,\dots,V$. To model the joint distribution of multi-view and multi-frame images in a unified space, our autoregressive model considers two folds of causality: \textit{scale} and \textit{time} (Fig.~\ref{fig:dual-causal}). 

For the scale causality, we start from a single-frame case. The distribution of multi-view images in the same frame is modeled in a joint manner, since multi-view images describe a unified 3D space at a specific time step as a whole. For clarity, we omit the timestep superscripts.
\begin{equation}
p(X_1^{1:V},\dots,X_K^{1:V})=\prod_{k=1}^Kp(X_k^{1:V}|X_1^{1:V},\dots,X_{k-1}^{1:V}).
    \label{eq:scale}
\end{equation}

For time causality, the generation of each frame is conditioned on historical information. Different from some prior works~\cite{wen2024panacea,russell2025gaia2controllablemultiviewgenerative}, we do not assume strong inter-dependency among multi-frame images from the same camera since this inductive bias cannot be maintained with complex ego-motions (\eg turning). In contrast, we condition the multi-view image generation for the current timestep on all the views from past frames.
\vspace{-5pt}
\begin{equation}
p(X_{1:K}^{1:V,1:T})=\prod_{t=1}^Tp(X_{1:K}^{1:V,t}|X_{1:K}^{1:V,1:t-1}).
    \label{eq:time}
\end{equation}

Combining the scale and time causality from Eq.~\ref{eq:scale} and Eq.~\ref{eq:time}, we formulate the dual-causal block in \ourmethod{} as
\begin{equation}
p\left(X_{1:K}^{1:V,1:T}\right)=\prod_{t=1}^T\prod_{k=1}^Kp\left(X_{k}^{1:V,t}|X_{1:k-1}^{1:V,1:t}\right).
    \label{eq:formulation}
\end{equation}
It models the joint distribution of multi-view and multi-frame images with both spatial and temporal consistency.

\subsection{Isotropic Spatio-Temporal Representation}
\label{sec:space}

Inductive biases, such as multi-view adjacency or 3D latent features, in previous work limit their generalization in open world and adaptivity to heterogeneous data. Alternatively, we propose an isotropic 4D position embedding by extending the relative rotary position embedding~\cite{su2024roformer} to the camera ray space shared by all the multi-view and multi-frame visual tokens to enhance spatial and temporal consistency.

To this end, we start from a token-wise Pl\"ucker ray~\cite{plucker1865xvii} for the multi-scale feature maps. For convenience, we define a global 3D coordinate space whose origin is the ego-vehicle position at the first timestep. The extrinsic parameters for each camera view $v$ at each timestep $t$ are denoted as $\mathbf{P^{v,t}}$, while the intrinsic parameter is denoted as $\mathbf{K}^{v,t}$.\footnote{Although the intrinsic parameters are commonly fixed along the temporal dimension, we keep the superscript $t$ to tolerate special varying cases.} For each visual token $\mathbf{x}^{v,t}_k$ at scale $k$, it is straightforward to compute the corresponding Pl\"ucker ray $\mathbf{p}^{v,t}_k$ based on the camera parameters and its location in the image space. We further abuse this notation by appending an extra timestep $t$, so the extended Pl\"ucker ray is written as 
\begin{equation}
    \mathbf{p}^{v,t}_k=(\mathbf{m}_k^{v,t}\in\mathbb{R}^3,\mathbf{d}_k^{v,t}\in\mathbb{R}^3,t)
    \label{eq:ray}
\end{equation}
where $\mathbf{m}_k^{v,t}=\mathbf{o}^{v,t}\times\mathbf{d}_k^{v,t}$, camera center $\mathbf{o}^{v,t}$ is in the global coordinate system, and unit vector $\mathbf{d}_k^{v,t}$ is the ray direction.

Despite an intuitive approach that attaches the encoded Pl\"ucker ray to the corresponding visual token~\cite{jin2024lvsm,chen2024unimlvg}, this absolute position embedding has limited extrapolation ability beyond the scene range in the training stage, thereby hindering the model's generalization. To this end, we introduce \textit{relative embedding} that conditions networks on the relative position between camera rays using the modified self-attention blocks, where the attention score logit between two visual tokens is calculated as:
\begin{equation}
    a_{i,j}=\left(\mathbf{R}_{k_i}^{v_i,t_i}\mathbf{q}_{k_i}^{v_i,t_i}\right)^T\left(\mathbf{R}_{k_j}^{v_j,t_j}\mathbf{k}_{k_j}^{v_j,t_j}\right)={\mathbf{q}_{k_i}^{v_i,t_i}}^T\mathbf{R}_{\Delta}^{i,j}\mathbf{k}_{k_j}^{v_j,t_j}
    \label{eq:rope}
\end{equation}
where $\mathbf{R}_{\Delta}^{i,j}={\mathbf{R}_{k_i}^{v_i,t_i}}^T\mathbf{R}_{k_j}^{v_j,t_j}$ reflects the relative position between visual token $i$ and $j$ in the camera ray space. We derive $\mathbf{R}_{k}^{v,t}$ by extending RoPE to 7D space based on the Pl\"ucker ray (Eq.~\ref{eq:ray}). For clarity, we omit the superscripts and subscripts for scale $k$, camera view $v$, and timestep $t$. The rotary position encoding is written as:
\begin{equation}
    \mathbf{R}=\left[\begin{matrix}
        \mathbf{R_m} & 0 & 0\\
        0 & \mathbf{R_d} & 0\\
        0 & 0 & \text{RoPE}_{\frac{d}{4}}(t)
    \end{matrix}\right]\\
\end{equation}
where $\text{RoPE}_{\frac{d}{4}}(t)\in\mathbb{R}^{\frac{d}{4}\times\frac{d}{4}}$ is the 1D rotary position embedding~\cite{su2024roformer} for time step and $d$ is head dimension in the multi-head self-attention module~\cite{vaswani2017attention}. $\mathbf{R_m}$ and $\mathbf{R_d}$ refer to the rotary position encoding for $\mathbf{m}$ and $\mathbf{d}$ terms in the Pl\"ucker ray. Given their small scales, we multiply a scalar factor to each of them separately.
\begin{equation}
\begin{aligned}
    \mathbf{R_m}&=\left[\begin{matrix}
        \text{RoPE}_{\frac{d}{8}}(\mathbf{m}_1) & 0 & 0\\
        0 & \text{RoPE}_{\frac{d}{8}}(\mathbf{m}_2) & 0\\
        0 & 0 & \text{RoPE}_{\frac{d}{8}}(\mathbf{m}_3)\\
    \end{matrix}\right]\\
    \mathbf{R_d}&=\left[\begin{matrix}
        \text{RoPE}_{\frac{d}{8}}(\mathbf{d}_1) & 0 & 0\\
        0 & \text{RoPE}_{\frac{d}{8}}(\mathbf{d}_2) & 0\\
        0 & 0 & \text{RoPE}_{\frac{d}{8}}(\mathbf{d}_3)\\
    \end{matrix}\right]\\ 
\end{aligned}
\end{equation}

This relative position embedding considers visual tokens from all scales, views, and frames in a unified 4D space with minimal inductive biases, enabling generalization over long ranges and horizons. The model can adaptively learn interactions between visual tokens across spatial and temporal domains, and accommodate heterogeneous camera parameters and frame rates in training and inference.

\subsection{Dual-Causal Block Architecture Design}
\label{sec:framework}

We build a transformer-based framework for high-quality multi-view video generation upon a pretrained ``next-scale prediction" image generation model~\cite{han2024infinity} with minimal 3D inductive bias. In each transformer block, visual tokens are sequentially passed through an \textit{image-wise self-attention module} for image realism, a \textit{global self-attention module} for spatio-temporal consistency, and an \textit{image-wise cross-attention module} for conditional fidelity, as shown in Fig.~\ref{fig:overview}.

\paragraph{Image-wise self-attention for image realism.} The image-wise attention follows the design in  Infinity~\cite{han2024infinity}. Each image is processed independently, conditioned on its prefix scales, regardless of frames and camera views. Axial 2D RoPE~\cite{heo2024rotary} on the image space is adopted, which is normalized to a specific resolution for multi-scale features.

\paragraph{Global self-attention for spatio-temporal consistency.} Previous methods~\cite{wang2024driving,wen2024panacea,russell2025gaia} often apply separate spatial and temporal attention to reduce computational complexity. However, this decoupled design includes some prior assumptions about the camera motion and scene dynamics. In contrast, we propose a unified global attention based on the formulation in Eq.~\ref{eq:formulation} where each visual token attends to all its prefix in both scale-wise and temporal topological orders from all the camera views (Fig.~\ref{fig:dual-causal}). The relative position embedding explained in Sec.~\ref{sec:space} is also employed in this module to serve as a unified spatio-temporal representation. This module is implemented with masked self-attention during training, while the mask can be eliminated through proper KV-cache during inference. 

\paragraph{Image-wise condition alignment.} We apply an image-wise cross-attention module for input conditions, including text descriptions, 3D object bounding boxes, and HD maps. This module works on each image independently. The text descriptions are processed through a T5 encoder~\cite{raffel2020exploring}. For each object bounding box per frame, we project its eight corners to camera image space. It is possible that each box may be projected to one or more camera views. The projected coordinates of eight corners in the image space are encoded with MLP and then concatenated with the object category embedding from T5 encoder~\cite{raffel2020exploring} to obtain a high-dimensional latent feature. For the HD map, we sample several 3D points for each map element. In contrast to directly encoding the structural map element (\eg polylines or masks), this sampling approach is more flexible when dealing with unstructured or sketched maps. The points are projected to camera image space similar to object bounding box corners. The project coordinates of points belonging to the same map element are encoded with a lightweight PointNet-style network~\cite{qi2017pointnet}, and are then concatenated with the map element category embedding from the T5 encoder~\cite{raffel2020exploring} to obtain a high-dimensional feature. The latent features for text, object boxes, and HD maps are combined in the cross-attention module, allowing the visual tokens to attend to them simultaneously. For both objects and map conditions, we employ an Axial 2D RoPE~\cite{su2024roformer,heo2024rotary} between the features and tokens based on their projected centers in the image space. Experiments show that this relative position embedding can notably improve the locality of conditions.

Given the multi-view order-invariant nature of the above modules, we attach a frame-wise local absolute Pl\"ucker ray embedding in the beginning of our model to distinguish different views. The image-wise self-attention and cross-attention modules inherit the pretrained weights from Infinity~\cite{han2024infinity}, while the global self-attention module is randomly initialized and combined with zero-initialization~\cite{zhang2023adding}. This alternating local-global module can help balance the spatio-temporal consistency and per-image realism.

\begin{algorithm}[tb!]
    \caption{\textbf{Recurrent Training for Long-Horizon Video Generation}}
    \label{alg:recurrent}
    \KwInput{Number of video frames $T$}
    \KwInput{Maximal cache length $M$}
    \KwInput{Multi-view video generation model that returns latent feature cache $G_{\theta}^L$}
    \KwInput{KV projection $\text{Proj}_{\theta}^{KV}$ (included in $G_{\theta})$}
    
    \Loop{
        Initialize KV latent feature cache $\mathbf{L}_{KV}\leftarrow[]$\\
        \For{$t\in1,\dots,T$}{
            $\mathbf{\tilde{x}}^t=\text{Random\_Bitwise\_Error}(\mathbf{x}^t)$\\
            \algcomment{Add random noise in scale-wise features to simulate the prediction errors}
            
            $\mathbf{KV}=\text{Proj}_{\theta}^{KV}(\mathbf{L}_{KV})$
            $\mathbf{\hat{x}}^t,\mathbf{l}_{KV}^t\leftarrow G_{\theta}^L(\mathbf{\tilde{x}}^t;\mathbf{KV})$\\
            \If{$\text{length}(\mathbf{L}_{KV})>M$}{
                $\mathbf{L}_{KV}\text{.pop}(0)$\\
            }
            $\mathbf{L}_{KV}\text{.append}\left(\text{stop\_grad}(\mathbf{l}_{KV}^t)\right)$\\
            \algcomment{Cache latents $\mathbf{L}_{KV}$ instead of $\mathbf{KV}$ to obtain gradient for $\text{Proj}_{\theta}^{KV}$ in past frames}

            $loss^t=\text{Cross\_Entropy}(\mathbf{\hat{x}}^t,\mathbf{x}^t)$\\
            Accumulate gradient of $loss^t$\\
            \algcomment{Per-frame backpropogation to save VRAM}
        }
        Update parameter $\theta$\\
        Clear accumulated gradient\\
    }
    
\end{algorithm}

\subsection{Recurrent Training for Long-Horizon Videos}
\label{sec:training}

Given the GPU memory constraints, our autoregressive model training is limited to a few frames. Although it is extendable during inference through sliding windows, the generation of long-horizon videos suffers from distribution drift due to the gap between training and inference stages. 

To bridge this gap, we propose a novel training paradigm to align the training and inference distributions (Alg.~\ref{alg:recurrent}). It recurrently conducts forward and backward propagation frame by frame, and the gradients are accumulated until the end of each long sequence for model optimization. Obviously, the later frames are conditioned on the earlier frames. We cache the latent features in the global self-attention module since it is the only module operating across frames (Sec.~\ref{sec:framework}). While previous works often cache the keys and values separately~\cite{dai2019transformer,shoeybi2019megatron} during inference, we directly store the latent features before KV projection layer instead. In this case, we can 1) save half of the GPU memory and 2) ensure the KV projection layer in the computational graph, which is critical for temporal information extraction.

However, there is an error accumulation issue during inference since the ``next-scale prediction" framework takes ground-truth visual tokens as inputs during training. To this end, we draw inspiration from Infinity~\cite{han2024infinity} to randomly incorporate some mistakes in the visual token inputs during training to simulate the prediction errors during inference.

\section{Experiments}

\begin{table*}[t!]
\begin{minipage}[t]{\linewidth}
\caption{Multi-view Video Generation Performance.}
\label{tab:multiview}
\vspace{-10pt}
\centering
\begin{tabular}{lccccc}
\toprule
\multicolumn{1}{l}{\multirow{2}{*}{\textbf{Method}}} & \multirow{2}{*}{\textbf{Resolution}}& \multicolumn{3}{c}{\textbf{Metrics}}\\
& & FID$\downarrow$ & FVD$\downarrow$&Throughput$\uparrow$\\
\midrule
MagicDrive~\cite{gao2023magicdrive}& 224$\times$400&  16.2 & - & 1.76\\
X-Drive~\cite{xie2024x} &  224$\times$400 &  16.0 & - & 0.83\\
DriveDreamer~\cite{wang2023drivedreamer} & 256$\times$448& 14.9 & 341& 0.37\\
BEVWorld~\cite{zhang2024bevworld} & - & 19.0 & 154 & -\\
Panacea~\cite{wen2024panacea} & 256$\times$512 & 17.0 & 139 & 0.67\\
DrivingDiffusion~\cite{li2024drivingdiffusion} & 512$\times$512 & 15.8 & 332 & -\\
\midrule
\ourmethod{} & 384$\times$672 & \textbf{10.5}& \textbf{91} & \textbf{1.96}\\
\bottomrule
\end{tabular}
\end{minipage}
\vspace{-10pt}
\end{table*}

\subsection{Experimental Setups}
\label{sec:setups}

\noindent\textbf{Datasets.} \ourmethod{} can scale up to heterogeneous training data with varying camera configurations, resolutions, or frame rates. We combine two public datasets (nuScenes~\cite{caesar2020nuscenes} and nuPlan~\cite{caesar2022nuplanclosedloopmlbasedplanning}), which are converted to a unified ScenarioNet~\cite{li2023scenarionet} format. NuScenes~\cite{caesar2020nuscenes} dataset contains $\sim$5 hours of driving data collected by 6 cameras with bounding box and map annotations, and scene descriptions from OmniDrive~\cite{wang2025omnidrive}. For nuPlan~\cite{caesar2022nuplanclosedloopmlbasedplanning} dataset, we filter the entire dataset based on complete sensor coverage and extract $\sim$55 hours of data from 8 cameras with auto-labeled bounding boxes and maps. We apply GPT-4o mini~\cite{team2024gpt} for scene descriptions. \ourmethod{} can be further scaled to larger-scale data, with details in the supplementary materials.

\noindent\textbf{Evaluation Metrics.}  To evaluate synthetic image quality, we adopt Frechet Video Distance (FVD)~\cite{unterthiner2018towards} and Frechet Inception Distance (FID)~\cite{heusel2017gans}. We also evaluate fidelity to conditions through perception models using NDS for objects and mIoU for map. Unless otherwise specified, we conduct the evaluation on the nuScenes validation set~\cite{caesar2020nuscenes}.

\noindent\textbf{Implementation Details.} We build \ourmethod{} on two variants of the pretrained Infinity~\cite{han2024infinity} model with 130M and 2B parameters respectively. Our training pipeline includes three stages: 1) We start from training the model on low-resolution ($192\times336$) multi-view short clips 2) The model is further finetuned on short clips of high-resolution ($384\times672$) multi-view images. 3) Finally, we adopt the recurrent training in Sec.~\ref{sec:training} on long sequence of driving scenarios to learn the long-term temporal coherence. All the training is conducted on 32 NVIDIA A100 GPUs.

\subsection{Video Generation Performance}

\noindent\textbf{Results and Comparisons.} In Tab.~\ref{tab:multiview}, we compare the multi-view generation performance of \ourmethod{} with other existing world models. We do not include in-house data in our model training, which makes our training data scale less than most of other approaches. Results show that we attain FID and FVD much better than all the baselines, demonstrating better image quality and temporal coherence. Notably, \ourmethod{} delivers a throughput of \textbf{1.96 images/s}, much faster than diffusion baselines, demonstrating the efficiency of our autoregressive architecture. It is also worth mentioning that our model applies an image-based VAE encoder-decoder to minimize the inductive bias relevant to camera setups and ego-motions, although this design choice may hurt the FID and FVD metric. We also provide some qualitative visualizations in Fig. \ref{fig:teaser}. It can be observed that our model can generate under different conditions. Using dual-causal autoregression, \ourmethod{} not only matches or exceeds the performance of models trained on orders of magnitude more private data, but also operates at higher frame rates.

\noindent\textbf{Object \& Map Conditions.} Our model takes optional object and map inputs as extra local geometric conditions to control the multi-view video generation. In addition to qualitative visualization in Fig.~\ref{fig:teaser}, we provide some additional quantitative results of object and map condition fidelity. We apply some state-of-the-art perception models~\cite{wang2023exploring,xie2023sparsefusion,liu2022bevfusion} to our synthetic images in nuScenes validation set. Both vision-based and multi-modal (combined with ground-truth point clouds) perception models are considered in the evaluation to comprehensively evaluate both the semantic and geometric realism. It is worth mentioning that StreamPETR takes long sequences as input, requiring  not only object realism in each single frame but also temporal coherence over long sequences. In Tab.~\ref{tab:object_condition} and \ref{tab:map_condition}, our synthetic images show great fidelity to both objects and maps notably better than baselines. For map condition, the projection of coordinates to camera space is not perfect due to the lack of height information in the map annotation, but \ourmethod{} still shows robust performance. For object condition, our synthetic images can reach a performance similar to real images.

\begin{table}[tb]
        \caption{Fidelity to Object Condition}
        \label{tab:object_condition}
        \vspace{-10pt}
        \begin{subtable}[t]{0.48\linewidth}
            \centering
            \subcaption{Camera (StreamPETR\cite{wang2023exploring})}
            \label{tab:object_camera} 
            \resizebox{\linewidth}{!}{
            \begin{tabular}{c|c}
                \toprule
                \textbf{Method} & \textbf{NDS$\uparrow$}  \\
                \midrule
                \textcolor{gray}{Oracle} & \textcolor{gray}{46.9} \\
                \midrule
                Panacea~\cite{wen2024panacea} & 32.1 \textcolor{red}{(68\%)}\\
                \ourmethod{} & \textbf{41.9} \textcolor{darkgreen}{(89\%)}\\
                \bottomrule
            \end{tabular}
        }
        \end{subtable}
        \begin{subtable}[t]{0.48\linewidth}
            \centering
            \subcaption{Multisensor (SparseFusion\cite{xie2023sparsefusion})}
            \label{tab:object_fusion} 
            \resizebox{\linewidth}{!}{
            \begin{tabular}{c|c}
                \toprule
                \textbf{Method} & \textbf{NDS$\uparrow$}  \\
                \midrule
                \textcolor{gray}{Oracle} & \textcolor{gray}{72.8} \\
                \midrule
                X-Drive~\cite{xie2024x} & 69.6 \textcolor{red}{(95\%)}\\
                \ourmethod{} & \textbf{72.0} \textcolor{darkgreen}{(99\%)}\\
                \bottomrule
            \end{tabular}
            }
        \end{subtable}
\end{table}

\begin{table}[tb]
        \vspace{-10pt}
        \caption{Fidelity to Map Condition}
        \label{tab:map_condition}
        \vspace{-10pt}
        \begin{subtable}[t]{0.48\linewidth}
            \centering
            \subcaption{Camera (BEVFusion-C\cite{liu2022bevfusion})}
            \label{tab:map_camera} 
            \resizebox{\linewidth}{!}{
            \begin{tabular}{c|c}
                \toprule
                \textbf{Method} & \textbf{mIoU$\uparrow$}  \\
                \midrule
                \textcolor{gray}{Oracle} & \textcolor{gray}{57.1} \\
                \midrule
                MagicDrive~\cite{gao2023magicdrive} & 28.9 \textcolor{red}{(51\%)}\\
                \ourmethod{} & \textbf{34.9} \textcolor{darkgreen}{(61\%)}\\
                \bottomrule
            \end{tabular}
        }
        \end{subtable}
        \begin{subtable}[t]{0.48\linewidth}
            \centering
            \subcaption{Multisensor (BEVFusion\cite{liu2022bevfusion})}
            \label{tab:map_fusion} 
            \resizebox{\linewidth}{!}{
            \begin{tabular}{c|c}
                \toprule
                \textbf{Method} & \textbf{mIoU$\uparrow$}  \\
                \midrule
                \textcolor{gray}{Oracle} & \textcolor{gray}{63.0} \\
                \midrule
                MagicDrive~\cite{gao2023magicdrive} & 47.0 \textcolor{red}{(75\%)}\\
                \ourmethod{} & \textbf{49.9} \textcolor{darkgreen}{(79\%)}\\
                \bottomrule
            \end{tabular}
            }
        \end{subtable}
    \vspace{-10pt}
\end{table}

\begin{table}[tb!]
    \centering
    \caption{Motion Cues in Generated Videos}
    \label{tab:motion}
    \vspace{-10pt}
    \resizebox{0.9\linewidth}{!}{
    \begin{tabular}{c|ccc}
        \toprule
       \multirow{2}{*}{\textbf{Method}}  & \multicolumn{3}{c}{\textbf{Motion Planning} ($L_2\downarrow$)}  \\
         & $1s$ & $2s$ & $3s$\\
        \midrule
        \textcolor{gray}{Oracle} & \textcolor{gray}{0.41} & \textcolor{gray}{0.70} & \textcolor{gray}{1.05}\\
        \midrule
        \ourmethod{} & \textbf{0.42} \textcolor{darkgreen}{(98\%)} & \textbf{0.73} \textcolor{darkgreen}{(96\%)} & \textbf{1.09} \textcolor{darkgreen}{(96\%)}\\
        \bottomrule
    \end{tabular}
    }
\end{table}

\begin{table*}[htb]
    \centering
    \caption{Novel View Synthesis Performance with Camera Shifts.}
    \label{tab:nvs}
    \vspace{-10pt}
    \begin{tabular}{l|cc|cc|cc}
        \toprule
        \multirow{3}{*}{Method} & \multicolumn{2}{c|}{Shift $1m$} & 
        \multicolumn{2}{c|}{Shift $2m$} & \multicolumn{2}{c}{Shift $4m$}\\
        \cmidrule(lr){2-3} \cmidrule(lr){4-5} \cmidrule(lr){6-7}
        & FID$\downarrow$ & FVD$\downarrow$ & FID$\downarrow$ & FVD$\downarrow$ &FID$\downarrow$ &FVD$\downarrow$\\
        \midrule
        StreetGaussian~\cite{yan2024street} & 32.12  & 153.45  &  43.24 & 256.91 & 67.44 & 429.98 \\
        OmniRe~\cite{chen2024omnire}   & 31.48  & 152.01  & 43.31  & 254.52 & 67.36 & 428.20 \\
        FreeVS~\cite{wang2024freevs}   &  51.26  &  431.99  &  62.04  &  497.37  &  77.14  &  556.14 \\
        \midrule
        \ourmethod{} & \textbf{14.11}  & \textbf{117.20} & \textbf{15.58} & \textbf{122.60} & \textbf{17.48} & \textbf{143.48}\\
        \bottomrule
    \end{tabular}
\end{table*}

\noindent\textbf{Motion Cues in Synthetic Videos.} In addition to visual quality and condition fidelity, we further evaluate whether our synthetic videos support plausible driving decisions by applying an end-to-end planner (VAD~\cite{jiang2023vad}) pretrained on real scenes to our generated videos. Tab.~\ref{tab:motion} shows that the pretrained planner derives actions from synthetic videos consistent with real scenes. This reflects not only visual realism but also physical and motion plausibility in our generated videos. 

\noindent\textbf{Zero-shot to Unseen Camera Configurations.} The ray-level relative position embedding allows \ourmethod{} to generalize to unseen camera configurations in a zero-shot manner. An example of sensor setups different from nuScenes and nuPlan is visualized in Fig.~\ref{fig:waymo} of the appendix.

\subsection{Novel View Synthesis}
\label{sec:nvs}
The relative position embedding in camera ray space allows our model to synthesize images for novel views. We follow \cite{wang2024freevs} to evaluate the performance of \ourmethod{} for novel view synthesis task on nuScenes validation set. Different shifts $\{1m,2m,4m\}$ are applied to the camera viewpoints. We measure the FID and FVD metrics between the synthetic shifted multi-view videos and the original real videos. In Tab.~\ref{tab:nvs}, our FID and FVD metrics only drop slightly after viewpoint shift. Compared to baselines, \ourmethod{} exhibits a significant performance gain even if it does not include any 3D inductive bias such as radiance field or point clouds. These results demonstrate the great capability of \ourmethod{} in camera controllability and novel view synthesis in a geometry-free data-driven manner.

\begin{table}[tb]
    \centering
    \vspace{-10pt}
    \caption{Ablation Study on Camera Ray.}
    \label{tab:position}
    \vspace{-10pt}
    \begin{tabular}{cc|cc}
    \toprule
        \textbf{S.-T. Module} & \textbf{Pos. Emb.} & \textbf{FID$\downarrow$} & \textbf{FVD$\downarrow$}  \\
        \midrule
         \XSolidBrush & - &  18.7 & 214 \\
         \Checkmark & Absolute & \textbf{17.2} & 138 \\
         \Checkmark & Relative & \textbf{17.2} & \textbf{124} \\
    \bottomrule
    \end{tabular}
\end{table}

\begin{figure}[tb!]
    \centering
    \includegraphics[width=\linewidth]{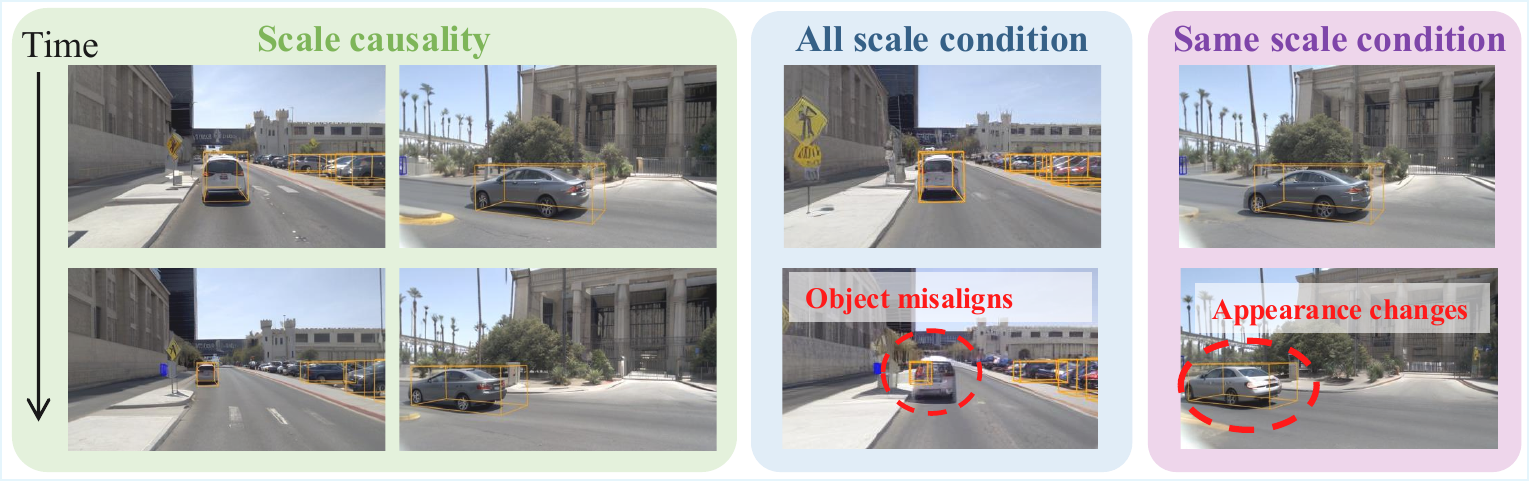}
    \vspace{-20pt}
    \caption{Ablation Study on Scale Causality. Conditioning on all scales in history hurts the modeling of dynamics, while conditioning only on same scale is insufficient for temporal coherence.}
    \label{fig:casuality}
    \vspace{-5pt}
\end{figure}

\begin{table}[tb!]
    \centering
    \caption{Ablation Study on Scale Causality and Model Size.}
    \label{tab:casuality}
    \vspace{-10pt}
    \begin{tabular}{cccc}
        \toprule
         \textbf{Past-frame Condition} & \textbf{Size} & \textbf{FID$\downarrow$} & \textbf{FVD$\downarrow$}  \\
        \midrule
         same scale & 130M & 20.7 & 151 \\
         all scales & 130M &  60.4 & 454 \\
         \midrule
         prefix scales & 130M & 17.2 & 124 \\
         \midrule
         prefix scales & 2B & \textbf{10.5} & \textbf{91}\\
        \bottomrule
    \end{tabular}
    \vspace{-15pt}
\end{table}

\begin{figure}[tb]
    \centering
    \includegraphics[width=\linewidth]{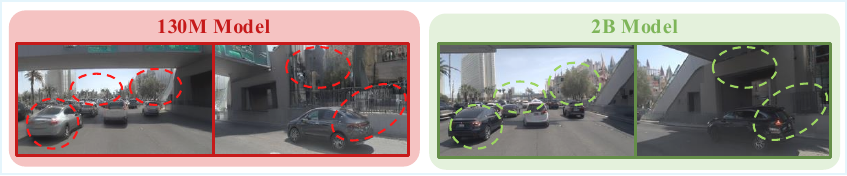}
    \vspace{-20pt}
    \caption{Ablation Study on Model Size. Large scale model can bring significantly better visual quality.}
    \label{fig:model_scale}
    \vspace{-10pt}
\end{figure}

\subsection{Ablation Study}
\noindent\textbf{Scale Causality.}
\ourmethod{} is built up on scale and time dual-causality. The temporal causality is 
inevitable if we would like to extend to long video autoregression. For scale causality (Eq.~\ref{eq:formulation}), different from low-scale condition in previous frames, we consider two alternatives conditions: 1) all scales or 2) same scale features in previous frames. Results are shown in Tab.~\ref{tab:casuality} and Fig.~\ref{fig:casuality}. Higher scales features of previous time steps cause future frames simply copying the details without simulating the dynamics, and same scale condition is far from enough for stable spatio-temporal modeling.

\noindent\textbf{Relative Ray Embedding.}
One of our key contributions lies in the relative position encoding in the camera ray space, which is critical for the spatio-temporal consistency. To justify our design, we consider several alternatives in Tab.~\ref{tab:position}. We find that this relative position embedding can perform better than absolute Pl\"ucker ray embedding, let alone none spatio-temporal reliance.

\noindent\textbf{Recurrent Training.} The recurrent training stage is greatly helpful for temporal coherence in video generation. It improves FVD from 100 to \textbf{91} within hundreds of iterations.

\noindent\textbf{Model Size.} Our minimal reliance on inductive bias provides a data-driven framework. Large model can exhibit significantly superior performance. We compare two variants of our model with 130M and 2.8B parameters. In Fig.~\ref{fig:model_scale} and Tab.~\ref{tab:casuality}, with the same training data, larger model enables much better image quality.

\section{Conclusion}

This paper introduces \ourmethod{}, a versatile autoregressive 4D world foundation model. Built from dual-causal blocks, \ourmethod{} performs unified 4D spatio-temporal reasoning by autoregressing along both the \textit{scale} and \textit{time} topological orders. To enhance spatio-temporal consistency, we design a novel isotropic relative position embedding in the continuous 7D Pl\"ucker ray space, which represents multi-view, multi-frame, and multi-scale visual features in a unified manner. To mitigate distribution drift in long-horizon video generation, we further adopt a recurrent training paradigm that better aligns the distributions of training and inference. Experiments demonstrate that \ourmethod{} can generate realistic multi-view videos with high visual fidelity, strong spatio-temporal coherence, and low latency under diverse conditioning signals, while generalizing to novel views and unseen camera configurations.

{
    \small
    \bibliographystyle{ieeenat_fullname}
    \bibliography{arxiv}
}

\appendix
\clearpage
In this supplementary material, we present additional implementation details and results that could not be included in the main paper due to space constraints. In Sec.~\ref{app:implementation}, we provide further details on the model architecture and training procedure. In Sec.~\ref{app:ablation}, we report additional ablation studies to support our design choices. In Sec.~\ref{app:results}, we include more qualitative results. Finally, we discuss the limitations of our work in Sec.~\ref{app:limitation}.

\section{Implementation Details}
\label{app:implementation}
As mentioned in Sec.~\ref{sec:setups}, \ourmethod{} model is initialized from the pretrained weights of Infinity~\cite{han2024infinity}.  Although video VAEs~\cite{openai2024sora,yang2025cogvideoxtexttovideodiffusionmodels} are widely recognized for improving temporal coherence and yielding better FVD scores, we instead directly adopt the multi-scale image tokenizer used in Infinity~\cite{han2024infinity} to better accommodate drastic camera motions and varying frame rates. 

\ourmethod{} includes two model variants with approximately 130M and 2B parameters. Unless otherwise specified, all experiments use the 2B model. The 130M model is only used in several ablation studies, including Tab.~\ref{tab:position}, Fig.~\ref{fig:model_scale}, and Tab.~\ref{tab:errors}.

The training of \ourmethod{} includes three stages as follows. In the first stage, we first train the model on short multi-view video clips containing 4 frames at a resolution of $192\times336$. The frame intervals are independently sampled between $0.1\text{s}$ and $1\text{s}$. We use a batch size of 64 and a learning rate of \textit{5e-5}, and train for $4\text{k}$ iterations. In the second stage, we then train the model on the same short clips but at a higher resolution of $384\times672$. The batch size is reduced to 32, and the learning rate is decreased to \textit{2.5e-5}. This stage also runs for $4\text{k}$ iterations. In the last stage, we adopt the recurrent training scheme described in Alg.~\ref{alg:recurrent}. The model is trained on long multi-view image sequences with 20 frames at $384\times672$ resolution. We use a batch size of 32 and a learning rate of \textit{2.5e-6} for 300 iterations. All training is conducted on 32 NVIDIA A100 GPUs using the AdamW optimizer~\cite{loshchilov2017decoupled}.

\begin{table}[tb]
    \centering
    \caption{Ablation Study on Spatio-Temporal Module.}
    \label{tab:global}
    \vspace{-10pt}
    \begin{tabular}{c|cc}
    \toprule
        \textbf{S.-T. Module} & \textbf{FID}$\downarrow$ & \textbf{FVD}$\downarrow$   \\
    \midrule
        decoupled & 15.6 & 140 \\
        global & \textbf{10.5} & \textbf{91} \\ 
    \bottomrule
    \end{tabular}
\end{table}

\begin{table}[tb]
    \centering
    \caption{Ablation Study on Random Errors in Training.}
    \label{tab:errors}
    \vspace{-10pt}
    \begin{tabular}{c|cc}
    \toprule
        \textbf{Random Errors} & \textbf{FID}$\downarrow$ & \textbf{FVD}$\downarrow$   \\
    \midrule
        \XSolidBrush & 19.8  & 142  \\
        \Checkmark & \textbf{17.2} & \textbf{124} \\ 
    \bottomrule
    \end{tabular}
\end{table}

\section{Additional Ablation Study}
\label{app:ablation}
\paragraph{Global Causal Attention.}
In Sec.~\ref{sec:framework}, we introduce a global self-attention module within the dual-causal block that jointly attends over all cameras and frames to ensure spatio-temporal consistency. Following prior work~\cite{gao2023magicdrive,wen2024panacea,wang2024driving}, we also experiment with an alternative decoupled design that applies two sequential self-attention modules: one for cross-view interactions per frame and one for cross-frame interactions per camera. As shown in Tab.~\ref{tab:global}, the global attention module significantly outperforms the decoupled spatio–temporal design. The decoupled approach implicitly assumes strong temporal correlation across frames captured by the same camera, thereby injecting strong ego-motion priors. In contrast, the global attention module incorporates minimal inductive bias and therefore generalizes better.

\paragraph{Random Errors in Training.}
\ourmethod{} inherits the teacher-forcing autoregressive training scheme used in VAR~\cite{tian2024visual} and Infinity~\cite{han2024infinity}, which can lead to error propagation due to the mismatch between training and inference. To mitigate this issue, we follow the idea introduced in Infinity~\cite{han2024infinity} and inject random errors into the bit-wise multi-scale tokens $X_k$ across all three training stages by randomly flipping a portion of bits in each token. As shown in Tab.~\ref{tab:errors}, this simple strategy improves video generation quality by narrowing the gap between training and inference.

\section{Additional Qualitative Results}
\label{app:results}

\begin{figure*}
    \centering
    \includegraphics[width=0.95\linewidth]{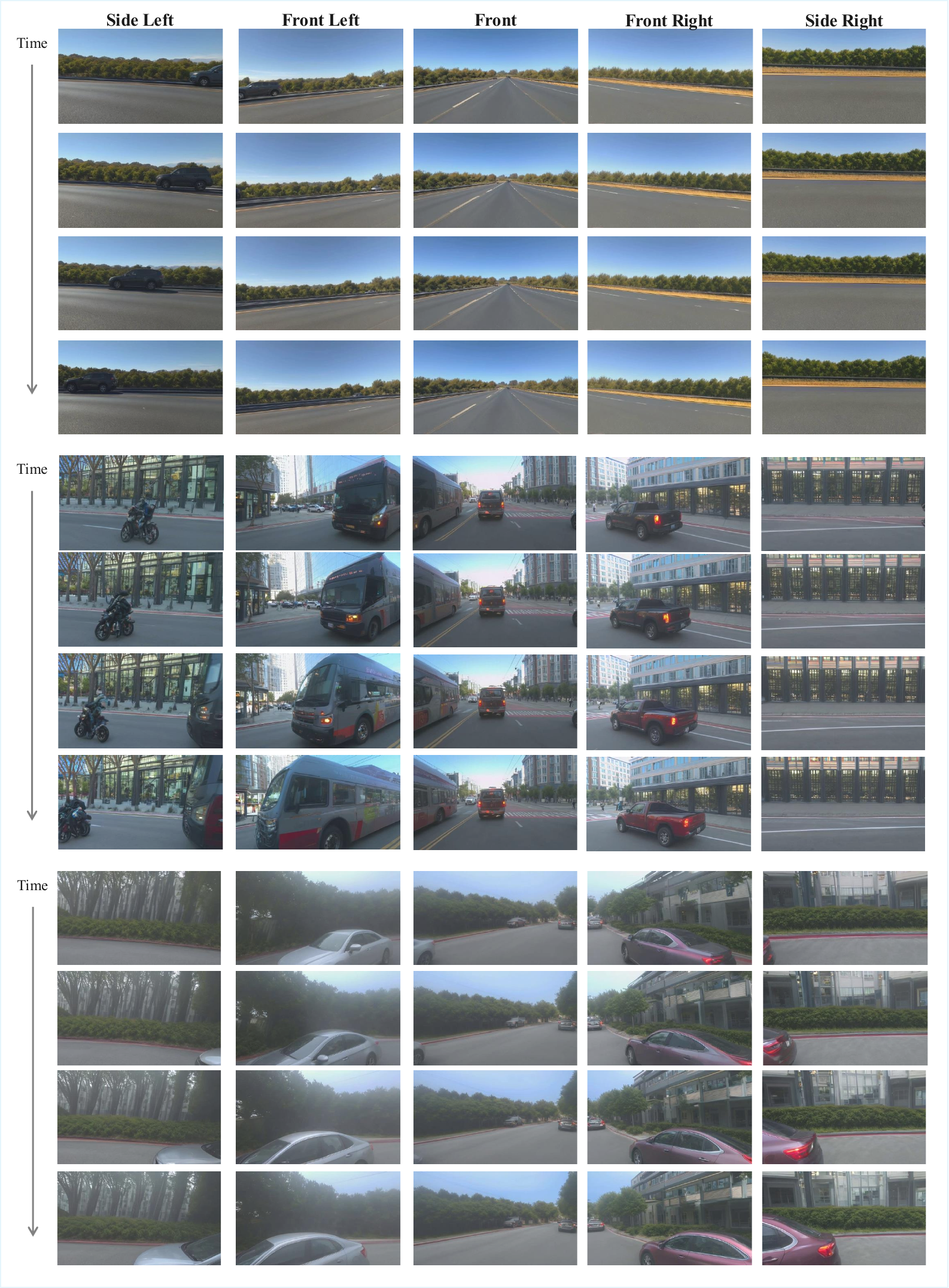}
    \vspace{-10pt}
    \caption{Zero-Shot to Unseen Waymo Open Dataset Camera Configuration.}
    \label{fig:waymo}
\end{figure*}

\paragraph{Zero-Shot to Novel Camera Configurations.} The ray-level relative position embedding enables \ourmethod{} to generalize to unseen camera configurations. In Fig.~\ref{fig:waymo}, we generate multi-view videos under the camera setup of the Waymo Open Dataset~\cite{sun2020scalability}, which is never seen during training. Nevertheless, \ourmethod{} is able to synthesize multi-view videos with reasonable visual quality and temporal coherence.

\begin{figure*}
    \centering
    \begin{subfigure}{0.9\linewidth}
        \includegraphics[width=\linewidth]{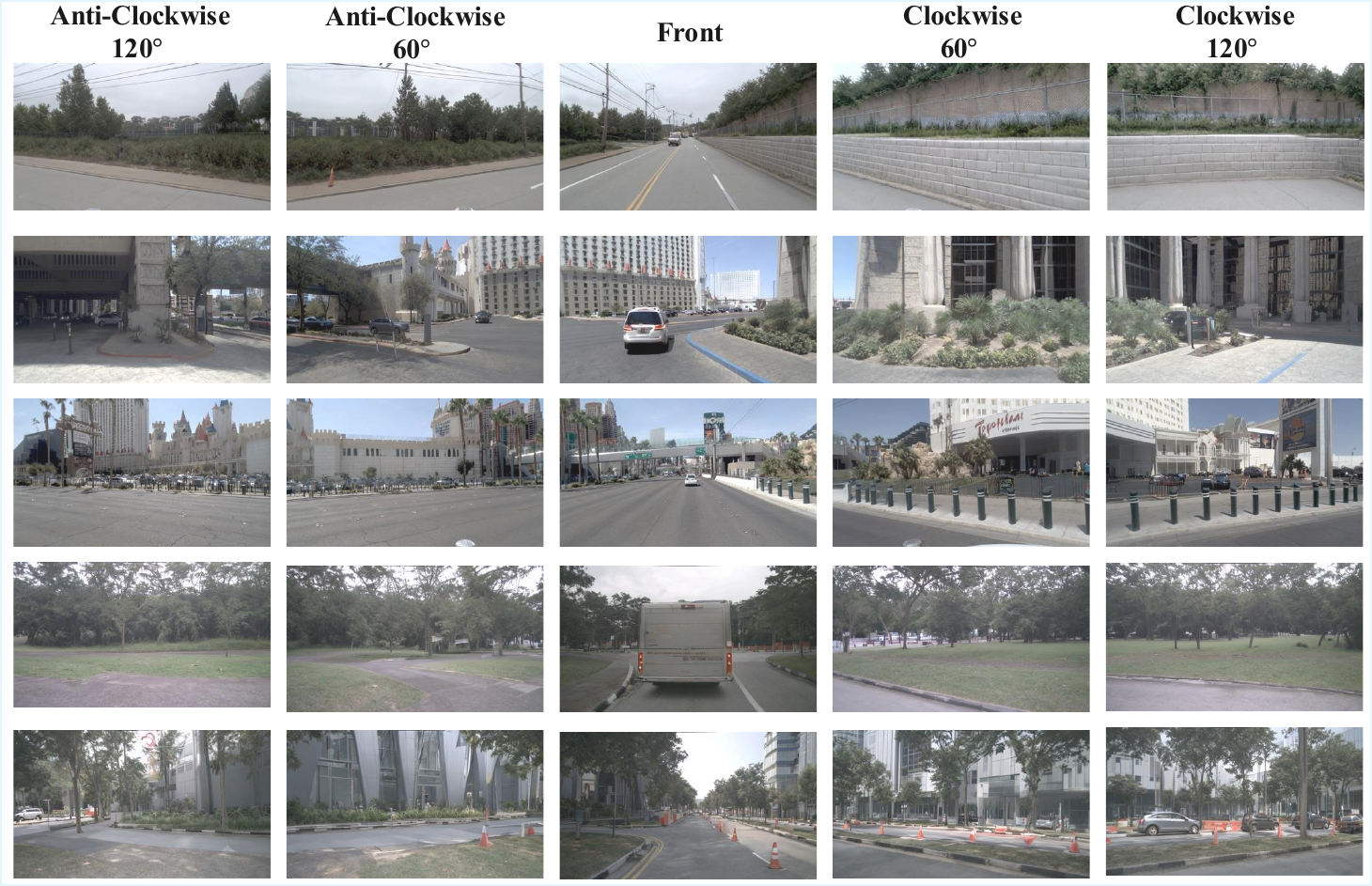}
        \caption{Camera Rotation.}
    \end{subfigure}
    \begin{subfigure}{0.9\linewidth}
        \includegraphics[width=\linewidth]{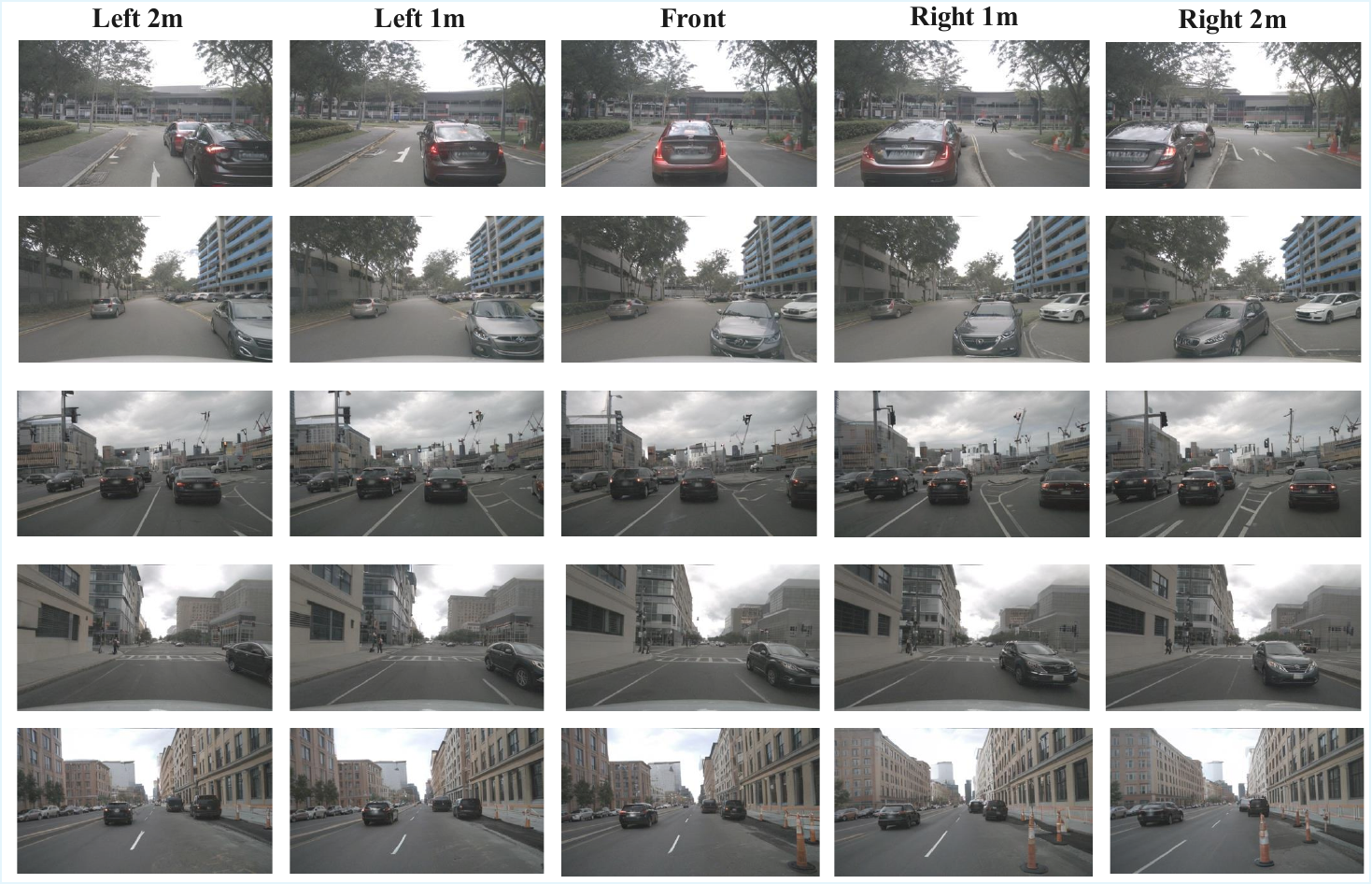}
        \caption{Camera Shifting.}
    \end{subfigure}
    \caption{Novel View Synthesis with Camera Rotation and Shifting.}
    \label{fig:app_camera}
\end{figure*}

\paragraph{Camera Rotation \& Shifting.} In Fig.~\ref{fig:app_camera}, \ourmethod{} is able to simulate camera rotations and translations. This highlights another advantage of our ray-level position embedding: it enables \ourmethod{} to effectively function as a feed-forward novel view synthesis model.

\begin{figure*}
    \centering
    \includegraphics[width=\linewidth]{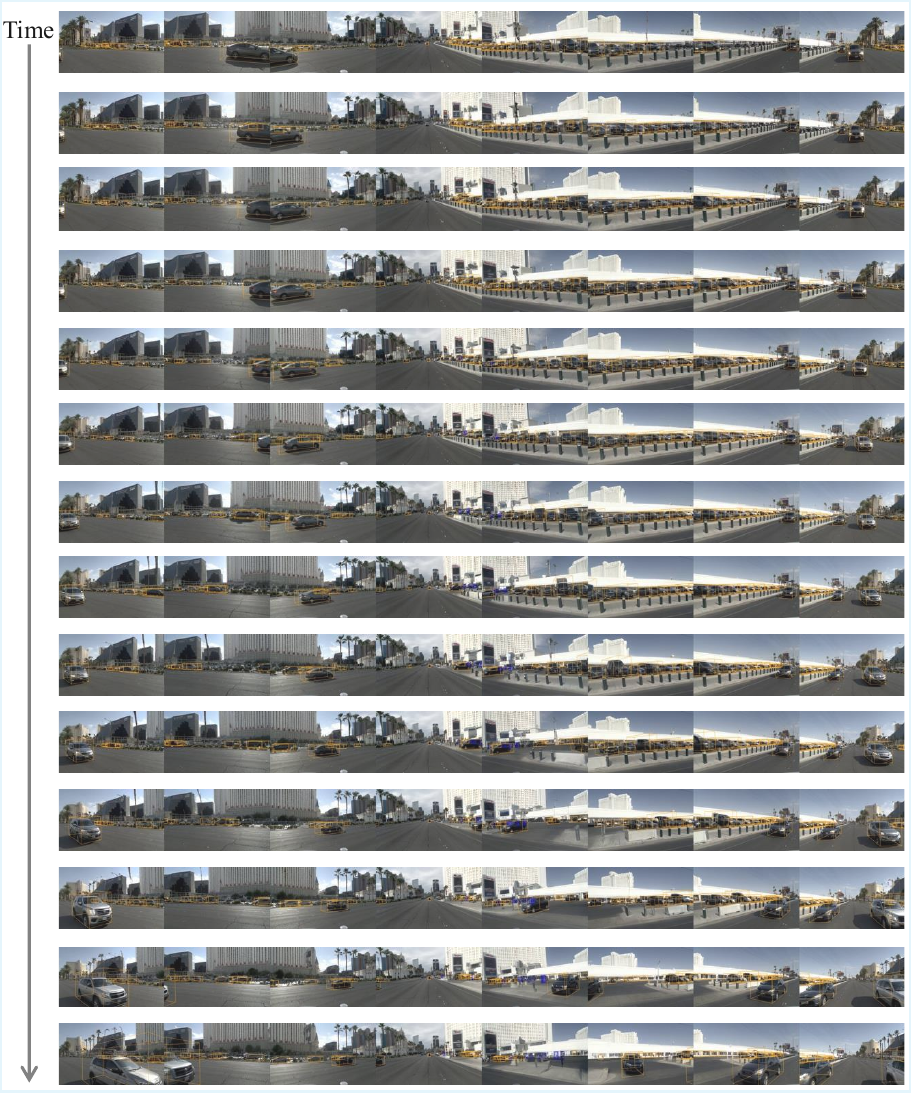}
    \caption{Multi-View Video Generation with Eight Cameras.}
    \label{fig:nuplan_video}
\end{figure*}

\begin{figure*}
    \centering
    \includegraphics[width=0.94\linewidth]{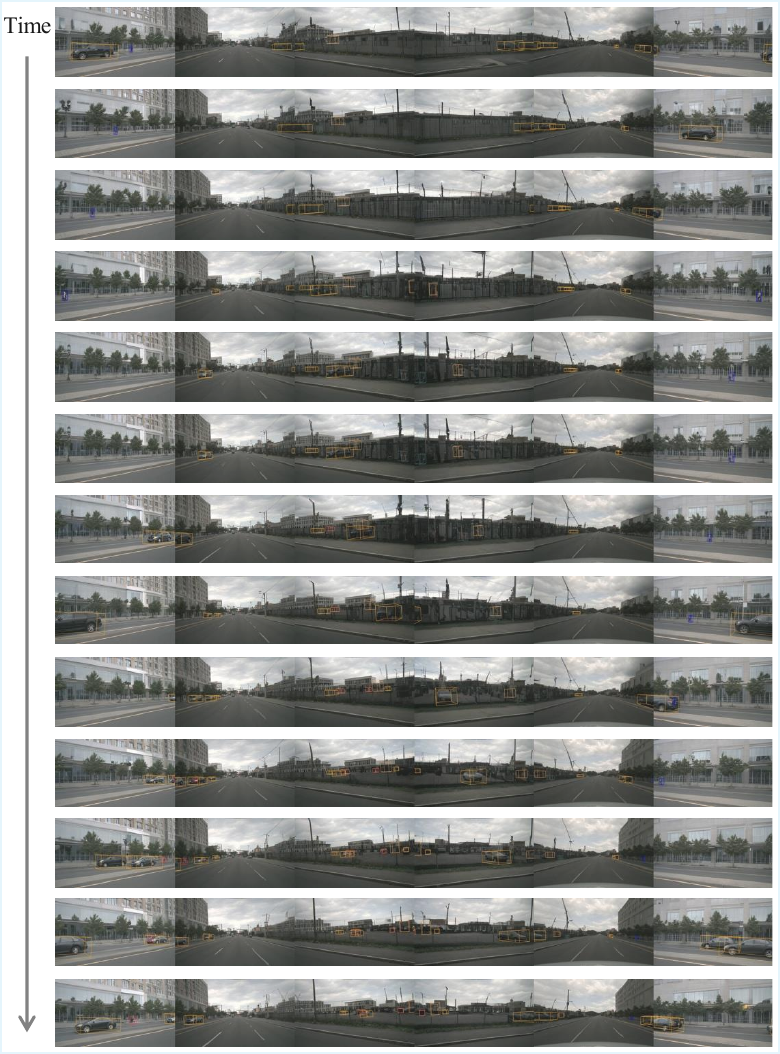}
    \caption{Multi-View Video Generation with Six Cameras.}
    \label{fig:nuscenes_video}
\end{figure*}

\paragraph{Multi-View Video Generation.} In Fig.~\ref{fig:nuplan_video} and Fig.~\ref{fig:nuscenes_video}, we present additional examples of multi-view video generation across diverse camera configurations and environments. We also provide several synthetic multi-view videos in the \textit{videos/} folder using different frame rates (with object and map conditions visualized). \ourmethod{} demonstrates strong condition fidelity and high visual realism.

\section{Limitation and Future Work}
\label{app:limitation}
Our current training data are limited to driving scenarios, which constrains the generalization ability of \ourmethod{} in non-driving environments. In future work, we plan to incorporate more diverse datasets (\eg, robotics or UAV) into the training pipeline. In addition, our evaluation primarily focuses on multi-view video generation. We aim to explore broader applications of \ourmethod{}, such as closed-loop simulation. Recent advances in visual autoregressive models~\cite{guo2025fastvar,voronov2024switti} can also be readily integrated into the \ourmethod{} framework to further enhance its capability and efficiency.

\newpage

\end{document}